\documentclass[runningheads]{llncs}

\usepackage{eccv}

\usepackage{eccvabbrv}

\usepackage{graphicx}
\usepackage{booktabs}

\usepackage[accsupp]{axessibility}  % Improves PDF readability for those with disabilities.

\usepackage{hyperref}

\usepackage{orcidlink}

\usepackage{enumitem}
\usepackage{balance}

\usepackage{ulem}
\usepackage{todonotes}
\usepackage[T1]{fontenc}
\usepackage{soul}
\usepackage{enumitem} % Include this package to customize lists
\usepackage[rightcaption]{sidecap}

\definecolor{cvprblue}{rgb}{0.21,0.49,0.74}
\definecolor{plt:green}{HTML}{2ca02c}
\definecolor{plt:red}{HTML}{d62728}
\definecolor{plt:blue}{HTML}{1f77b4}
\definecolor{plt:orange}{HTML}{ff7f0e}

\newcommand{\parag}[1]{\smallskip\noindent\textbf{#1}\enspace}

\begin{document}

%Reliability in Semantic Segmentation: \\Can We Use Synthetic Data?
\title{The BRAVO Semantic Segmentation Challenge Results in UNCV2024} 

\titlerunning{The BRAVO Challenge 2024}

\author{
Tuan-Hung Vu\inst{1} \and
Eduardo Valle\inst{1} \and
Andrei Bursuc\inst{1} \and
Tommie Kerssies\inst{2} \and 
Daan de Geus\inst{2} \and 
Gijs Dubbelman\inst{2} \and
Long Qian\inst{3,4} \and 
Bingke Zhu\inst{3,5} \and 
Yingying Chen\inst{3,5} \and 
Ming Tang\inst{3,4} \and 
Jinqiao Wang\inst{3,4,5} \and
Tomáš Vojíř\inst{6} \and 
Jan Šochman\inst{6} \and
Jiří Matas\inst{6} \and
Michael Smith\inst{7} \and
Frank Ferrie\inst{7} \and
Shamik Basu\inst{8} \and
Christos Sakaridis\inst{10} \and\\
Luc Van Gool\inst{9,10}
}

\authorrunning{T-H.Vu et al.}

\institute{valeo.ai, France (Challenge Organizers) \and
Eindhoven University of Technology, Netherlands \and
Chinese Academy of Sciences, China \and
University of Chinese Academy of Sciences, China \and
Objecteye Inc., China \and 
Czech Technical University in Prague, Czechia \and
McGill University, Canada \and
University of Bologna, Italy \and
Institute for Computer Science, Artificial Intelligence and Technology, Bulgaria \and
ETH Zurich, Switzerland
}

\maketitle
\begin{abstract}
    We propose the unified BRAVO challenge to benchmark the reliability of semantic segmentation models under realistic perturbations and unknown out-of-distribution (OOD) scenarios. We define two categories of reliability: (1) semantic reliability, which reflects the model's accuracy and calibration when exposed to various perturbations; and (2) OOD reliability, which measures the model's ability to detect object classes that are unknown during training. The challenge attracted nearly 100 submissions from international teams representing notable research institutions. The results reveal interesting insights into the importance of large-scale pre-training and minimal architectural design in developing robust and reliable semantic segmentation models.
\end{abstract}

\section{BRAVO Challenge}
Autonomous vehicles are safety-critical systems operating in a complex open world. As such, they must not only deliver excellent performance in their operational design domain but also be provably robust to adversarial attacks, extreme weather conditions, domain changes, and rare but potentially catastrophic driving situations. The BRAVO Challenge aims to develop test beds to assess and statistically demonstrate the robustness of driving perception models. The challenge employs existing test sets, sometimes with added synthetic augmentations, with novel test metrics to emphasize safety-centered challenges: calibration of models' outputs and estimation of their uncertainty; detection of out-of-
% domain
distribution inputs, at scene or object level; assessment of domain shifts. The BRAVO Challenge 2024 focused on semantic segmentation and was presented at the UNCV workshop~\footnote{\url{https://uncertainty-cv.github.io/2024/}}.

\subsection{Main Tracks}
In the BRAVO Challenge 2024, we proposed two tracks:
\begin{itemize}[leftmargin=*, label=--]
    \item \textbf{Track 1: single-domain Training.} Participants must train their models exclusively on the Cityscapes~\cite{cordts2016cityscapes} dataset. This track evaluates the robustness of models trained with limited supervision and geographical diversity when facing unexpected corruptions observed in real-world scenarios.
    \item \textbf{Track 2: multi-domain Training.} Participants may train their models over a mix of datasets, whose choice is strictly limited to the list provided below, comprising both natural and synthetic domains. This track assesses the impact of fewer constraints on the training data on robustness. The accepted datasets are: Cityscapes~\cite{cordts2016cityscapes}, BDD100K~\cite{yu2020bdd100k}, Mapillary Vistas~\cite{neuhold2017mapillary}, India Driving Dataset~\cite{varma2019idd}, WildDash 2~\cite{zendel2022unifying}, GTA5~\cite{richter2016playing} and SHIFT~\cite{sun2022shift}.
\end{itemize}

\subsection{BRAVO Dataset}
\label{sec:bravo_dataset}
The BRAVO Challenge 2024 aimed to benchmark semantic segmentation models on urban scenes undergoing diverse forms of natural degradation and realistic-looking synthetic corruptions. To this end, we repurposed existing datasets~\cite{sakaridis2021acdc,chan2021segmentmeifyoucan,franchi2021robust} and combined them with newly generated data. The BRAVO Dataset  2024 comprised images from ACDC~\cite{sakaridis2021acdc}, SegmentMeIfYouCan (SMIYC)~\cite{chan2021segmentmeifyoucan}, Out-of-context Cityscapes~\cite{franchi2021robust}, and new synthetic data. We organized the dataset into six subsets, two with real data and four based on the validation set of Cityscapes with synthetic augmentations: 

\begin{itemize}[leftmargin=*, label=--]
	\item \textit{bravo-ACDC}: real scenes captured in adverse weather conditions, \textit{i.e.}, fog, night, rain, and snow~\cite{sakaridis2021acdc}; 
	\item \textit{bravo-SMIYC}: real scenes featuring out-of-distribution (OOD) objects rarely encountered on the road~\cite{chan2021segmentmeifyoucan}; 
	\item \textit{bravo-synrain}: 500 augmented scenes with synthesized raindrops on the camera lens~\cite{pizzati2023physics};
	\item \textit{bravo-synobjs}: 656 augmented scenes with inpainted synthetic OOD objects from 26 classes~\cite{loiseau2024reliability};
	\item \textit{bravo-synflare}: 308 augmented scenes with synthesized light flares~\cite{wu2021train};
	\item \textit{bravo-outofcontext}: 329 augmented scenes with random backgrounds for road and sidewalk~\cite{franchi2021robust}.
\end{itemize}

\subsection{Metrics}
The BRAVO Challenge 2024 evaluated methods on various metrics to assess their performance in semantic segmentation and out-of-distribution (OOD) detection. The semantic metrics assessed the quality of the semantic segmentation predictions on both accuracy and calibration. The OOD metrics assess the model’s ability to detect whether the objects are OOD, \textit{i.e.}, to distinguish between \textit{known} classes seen during training \vs \textit{unknown} classes seen at test time. The BRAVO Index combines the semantic and OOD metrics to rank the models.

\parag{Semantic metrics} are computed on all subsets, except SMIYC, for valid pixels only.
Valid pixels are those not invalidated by extreme uncertainty, such as pixels obscured by the brightest areas of a flare or covered by an OOD object.
\begin{itemize}[leftmargin=*, label=--]
	\item Mean Intersection over Union (mIoU): Proportion of correctly labeled pixels among all pixels. Only semantic metric that does not rely on prediction confidence. Higher values indicate better segmentation accuracy.
	\item Expected Calibration Error (ECE): Difference between predicted confidence and actual accuracy. Lower values indicate better calibration.
	\item Area Under the ROC Curve (AUROC): Area Under the ROC Curve over the binary criterion of a pixel being accurate, ranked by the predicted confidence level for the pixel. Higher values indicate better calibration, as the confidence ranking matches the correctness of the pixels. 
	\item False Positive Rate at 95\% True Positive Rate (FPR@95): False positive rate when the true positive rate is 95\% in the ROC curve above. Lower values indicate better calibration at the tail of the confidence distribution: the ability to reject false positives even when we reach the most true positives.
    \item AUPR-Success: Area Under Precision-Recall curve, over the same data as the AUROC. Higher values indicate the ability of higher confidence to match correct pixels and, thus, better calibration.
    \item AUPR-Error: Uses the reversed data (pixel being inaccurate, ranked by $1-$confidence). Higher values indicate the ability of lower confidence to match incorrect pixels and, thus, better calibration. That tends to be stricter than AUPR-Success, since incorrect pixels tend to be rarer.
\end{itemize}

\parag{OOD metrics} are computed on the SMIYC and SynObjs subsets only for invalid pixels, \ie, those obscured by OOD objects. The OOD metrics are computed over the binary criterion of a pixel being invalid, ranked by the reversed predicted confidence level for the pixel, and include:
\begin{itemize}[leftmargin=*, label=--] 
	\item Area Under the ROC Curve (AUROC).
	\item False Positive Rate at 95\% True Positive Rate (FPR@95).
	\item Area Under the Precision-Recall Curve (AUPRC).
\end{itemize}

\parag{Aggregated metrics.} For model ranking, the metrics above are aggregated as follows:
\begin{itemize}[leftmargin=*, label=--]
	\item Semantic. The harmonic mean of all semantic metrics, with ECE and FPR@95 reversed (as $1-x$).
	\item OOD. The harmonic mean of all OOD metrics, with FPR@95 reversed.
	\item BRAVO Index: The harmonic mean of Semantic and OOD, used as the official ranking metric for the challenge.
\end{itemize}

\begin{figure}[t!]
  \centering
  \includegraphics[width=\textwidth]{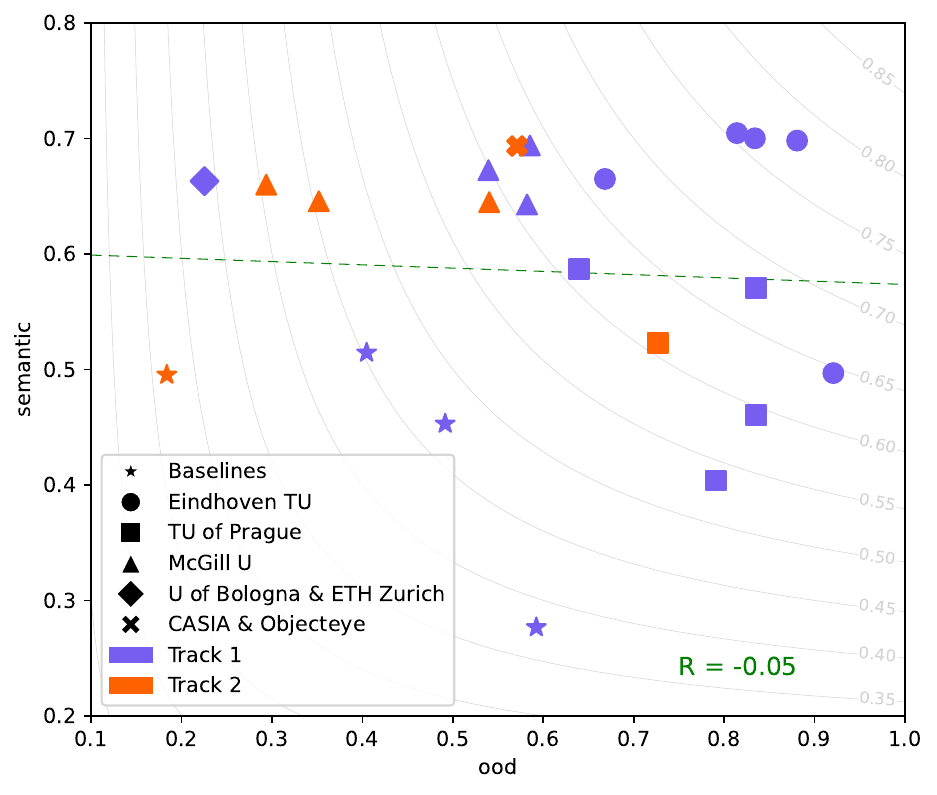}
  \caption{All submissions. Aggregated metrics (out-of-distribution and semantic) on axes, ranking metric (BRAVO Index) on level set. More freedom on the training dataset (Task 2, in orange) did not translate into better results.}
  \label{fig:example}
\end{figure}

In~\cref{sec:bravo_challenge_supp}, we provide more information on the rules and the submission format of the BRAVO challenge 2024.

\section{Submissions digest}
This section collects all the solutions from the two challenge tracks, along with interesting findings reported by the participants. We primarily retain the notation used by the participants in their reports. Notations are, thus, not consistent across subsections. The first person ``we'' in the submission subsections is that of the respective authors, summarized and sometimes paraphrased by the challenge organizers. Due to space limitations, the approach figures and some training details are provided in~\cref{sec:submission_digest_supp}.

\begin{table}[t!]
	\centering
	\resizebox{0.6\linewidth}{!}{
		\begin{tabular}{lcccc}
			\toprule
			\textbf{Method} & \textbf{Ref.} & \textbf{BRAVO$\uparrow$} & \textbf{Semantic$\uparrow$} & \textbf{OOD$\uparrow$} \\
			\midrule
            DINOv2-OOD & \cref{sec:dinov2ood} & \textbf{77.9} & \textbf{69.8} & \textbf{88.1} \\
			PixOOD w/ ResNet-101 DeepLabv3~\cite{vojivr2024pixood} & \cref{sec:pixelood1st}  & 61.2 & 58.7 & 64.0 \\
			Ensemble & \cref{sec:ensemble_track1} & 61.1 & 64.3 & 58.2 \\
            PhyFea~\cite{basu2024physically} & \cref{sec:phyfea} & 33.6 & 66.3 & 22.5 \\
            \midrule
            \textit{Baseline:} SegFormer-B5~\cite{xie2021segformer} & - & 47.1 & 45.3 & 49.2 \\
            \textit{Baseline:} ObsNet-ResNet101~\cite{besnier2021triggering} & - & 45.3 & 51.5 & 40.5 \\
			\textit{Baseline:} RbA Swin-B~\cite{nayal2023rba} & - & 37.7 & 27.7 & 59.2 \\
			\bottomrule
	\end{tabular}}
	\caption{Track 1 -- Performance metrics across different approaches.}
	\label{tbl:track1_all}
\end{table}

\begin{table}[t!]
	\centering
	\resizebox{0.52\linewidth}{!}{
		\begin{tabular}{lcccc}
			\toprule
			\textbf{Method} & \textbf{Ref.} & \textbf{BRAVO$\uparrow$} & \textbf{Semantic$\uparrow$} & \textbf{OOD$\uparrow$} \\
			\midrule
			InternImage-OOD & \cref{sec:internimage_ood} & \textbf{62.6} & \textbf{69.3} & 57.1 \\
			PixOOD~\cite{vojivr2024pixood}  & \cref{sec:pixelood2nd} & 60.8 & 52.3 & \textbf{72.7} \\
			Ensemble & \cref{sec:ensemble_track2} & 58.8 & 64.5 & 54.0 \\
            \midrule
            \textit{Baseline:} FAMix~\cite{fahes2024simple} & - & 26.8 & 49.5 & 18.4 \\
			\bottomrule
	\end{tabular}}
	\caption{Track 2 -- Performance metrics across different approaches.}
	\label{tbl:track2_all}
\end{table}

\subsection{Quantitative summary}

The results are summarized\footnote{Detailed results are available at the challenge server and at the repository: \url{https://github.com/valeoai/bravo_challenge}.} in \cref{tbl:track1_all,tbl:track2_all}, which show the best submission for each team. \cref{fig:example} shows all public submissions at the end of the Challenge.

\cref{fig:example} shows the submissions considerably improved over the proposed baselines. Due to the strictness of the harmonic mean, submissions that only enhanced one criterion were penalized on the ranking BRAVO Index. The most surprising collective finding is that multiple training datasets (Track 2) did not improve the metrics using only Cityscapes (Track 1). 

The OOD and semantic metrics were uncorrelated among the submissions (regression line in green in \cref{fig:example}, $R=-0.05$). On the other hand, we observed varying degrees of correlation among the metrics aggregated by the subsets listed in \cref{sec:bravo_dataset}. The correlogram appears in the~\cref{sec:submission_digest_supp}.

All top two solutions in both tracks leverage vision foundation models (VFMs) pretrained on massive data corpus, \textit{i.e.}, DINOv2~\cite{oquab2023dinov2} and InternImage~\cite{wang2022internimage}. For semantic segmentation, it was surprising to find that a simple linear decoder (DINOv2-OOD) outperformed more sophisticated decoders by a large margin, according to the unified BRAVO score.

Those findings emphasize the importance of robust VFM backbone and may temporarily shift researchers' focus toward training more robust VFMs, rather than concentrating on sophisticated network design for downstream tasks.
However, the current best BRAVO score is still far from perfect and using larger models is not showing clear beneficial signals, we believe that advancing reliable segmentation requires progress from both sides: developing more robust VFM backbones and designing more efficient architectures to better exploit the knowledge encapsulated in pre-trained VFMs.

\subsection{Track 1: DINOv2-OOD -- Eindhoven University of Technology}
\label{sec:dinov2ood}
\vspace{-5pt}
\textit{Authors: Tommie Kerssies, Daan de Geus, Gijs Dubbelman}
\vspace{3pt}

This solution fine-tunes pre-trained Vision Foundation Models (VFMs) for semantic segmentation, leveraging their robust representations.
Given a pre-trained VFM, we attach an off-the-shelf segmentation decoder and fine-tune the entire model for semantic segmentation.
We evaluate this meta-architecture in several different configurations. 
Our primary solution uses the DINOv2 VFM~\cite{oquab2023dinov2}, selected due to its effectiveness in domain generalized semantic segmentation for urban scenes~\cite{englert2024exploring,kerssies2024benchmark}.
DINOv2, built upon the Vision Transformer (ViT) architecture~\cite{dosovitskiy2020image}, is pre-trained using self-supervised learning on a vast, curated dataset. We experiment with all available sizes of DINOv2.

For our default segmentation decoder, we use a simple linear layer that transforms the patch-level features $\mathbf{F} \in \mathbb{R}^{E\times\frac{H}{P}\times\frac{W}{P}}$ into segmentation logits $\mathbf{L}\in\mathbb{R}^{C\times\frac{H}{P}\times\frac{W}{P}}$, where $H$ and $W$ represent the height and width of the input image, $P$ denotes the patch size, $E$ is the feature dimension, and $C$ is the number of classes in the dataset. We choose a linear layer trusting that the strong representations learned by the VFM forgo a more advanced decoder (which could also overfit to the training distribution).

We evaluate the impact of large-scale pre-training with DINOv2 by constrast with a DeiT-III~\cite{touvron2022deit} ViT pre-trained on ImageNet-1K\cite{deng2009imagenet} and fine-tuned on Cityscapes. We assess the impact of a more advanced decoder by contrast with a Mask2Former decoder~\cite{cheng2022masked}. We also contrast the default patch size ($16\times16$) to a more expensive $8\times8$.

\parag{Training.} 
When training the model with a linear decoder, we bilinearly upsample the segmentation logits $\mathbf{L}\in\mathbb{R}^{C\times\frac{H}{P}\times\frac{W}{P}}$ to $\mathbf{L'}\in\mathbb{R}^{C\times H\times W}$, and then apply a categorical cross-entropy loss to those logits and the semantic segmentation ground truth to fine-tune the model. When using Mask2Former, the decoder outputs a set of mask logits $\mathbf{M}\in\mathbb{R}^{N\times\frac{H}{P}\times\frac{W}{P}}$ and corresponding class logits $\textbf{C}\in\mathbb{R}^{N\times(C+1)}$, where N is the number of masks and C includes an additional ``no-object'' class. Following Mask2Former, during training those mask and class logits are matched to the ground truth using bipartite matching. The predicted masks are then supervised with a cross-entropy loss and a Dice loss, and the predicted classes are supervised with a categorical cross-entropy loss.

\parag{Testing.} During inference with the linear decoder, we compute per-pixel class confidence scores with softmax on the upsampled class logits $\mathbf{L'}$, using the highest score to predict the pixel class. For the Mask2Former decoder~\cite{cheng2022masked}, we bilinearly upsample the mask logits $\mathbf{M}\in\mathbb{R}^{N\times\frac{H}{P}\times\frac{W}{P}}$ to the original resolution, resulting in $\mathbf{M'}\in\mathbb{R}^{C\times H\times W}$. We then obtain the mask scores $\mathbf{P}_\textbf{M}=\texttt{sigmoid}(\mathbf{M'})$.

Overall per-pixel class confidence scores $\mathbf{P'} \in \mathbb{R}^{C \times H \times W}$ are computed by multiplying the mask scores with the class scores across all masks. Specifically, for each class $c$ and pixel $(h, w)$, we have: $\mathbf{P'}_{c,h,w} = \sum_{n=1}^{N} \mathbf{P}_{\mathbf{M_n},h,w} \cdot \mathbf{P}_{\mathbf{C_n},c}$
For each pixel, the predicted class is the one with the highest value in $\mathbf{P'}$, and we also output this maximum value as the confidence score.

\begin{table}[t!]
	\centering
	\resizebox{0.6\linewidth}{!}{
	\begin{tabular}{lccc}
		\toprule
		\textbf{Method} & \textbf{BRAVO $\uparrow$} & \textbf{Semantic $\uparrow$} & \textbf{OOD $\uparrow$} \\
		\midrule
		DINOv2, ViT-L, 8x8 patch size, linear decoder & \textbf{77.9} & 69.8 & 88.1 \\
		DINOv2, ViT-L, 16x16 patch size, linear decoder & 77.2 & \textbf{70.8} & 84.8 \\
		DINOv2, ViT-g, 16x16 patch size, linear decoder & 76.1 & 70.0 & 83.4 \\
		DINOv2, ViT-B, 16x16 patch size, linear decoder & 75.5 & 70.5 & 81.4 \\
		DINOv2, ViT-S, 16x16 patch size, linear decoder & 69.9 & 69.1 & 70.6 \\
		DINOv2, ViT-g, 16x16 patch size, Mask2Former decoder & 64.5 & 49.7 & \textbf{92.1} \\
		DeiT III (IN1K), ViT-S, 16x16 patch size, linear decoder & 54.1 & 62.8 & 47.6 \\
		\bottomrule
	\end{tabular}}
	
	\caption{Track 1 -- DINOv2-OOD -- Ablated models.}
	\label{tbl:dino_bravoindex}
\end{table}

\parag{Results.} As shown in \cref{tbl:dino_bravoindex}. Our best performing model, DINOv2 with a ViT-L/8 backbone and a linear decoder, achieves the highest BRAVO index of $77.9$, which is $+16.7$ higher than the next best method, \textit{i.e.}, PixelOOD presented in \cref{sec:pixelood1st}.

We observe in \cref{tbl:dino_semantic} that the model with the highest mIoU, DINOv2 with a ViT-g/16 backbone and a Mask2Former decoder, performs relatively poorly on the other metrics.
As the other metrics take into account the confidence score, this suggests that while Mask2Former is effective at predicting the correct class, it is less adept at estimating the confidence
of its predictions, at least in the out-of-the-box manner in which we used it.
In our setup, models with a simple linear decoder provide the best trade-off between segmentation accuracy and confidence estimation.

A similar result is observed when changing the patch size. A smaller patch size of 8 × 8 results in better mIoU, but the other metrics are worse. This indicates that a smaller patch size allows the model to capture more fine-grained details, which improves the accuracy of the predicted class labels, but that this somehow makes the confidence scores less reliable.

\begin{table}[t!]
	\centering
	\resizebox{0.8\linewidth}{!}{
	\begin{tabular}{lccccccc}
		\toprule
		\textbf{Method} & \textbf{mIoU $\uparrow$} & \textbf{AUPR-Error $\uparrow$} & \textbf{AUPR-Success $\uparrow$} & \textbf{AUROC $\uparrow$} & \textbf{ECE $\downarrow$} & \textbf{FPR@95 $\downarrow$} \\
		\midrule
		DINOv2, ViT-L, 8x8 patch size, linear decoder & {76.7} & 40.0 & 99.4 & 91.4 &  {2.0} & {38.8} \\
		DINOv2, ViT-L, 16x16 patch size, linear decoder & 75.9 & 41.2 & \textbf{99.5} & 92.3 & \textbf{1.7} & 37.8 \\
		DINOv2, ViT-g, 16x16 patch size, linear decoder & 77.6 & 39.3 & \textbf{99.5} & 92.3 & {1.8} & {37.6} \\
		DINOv2, ViT-B, 16x16 patch size, linear decoder & 71.7 & 43.3 & 99.4 & 92.3 & 2.1 & {40.3} \\
		DINOv2, ViT-S, 16x16 patch size, linear decoder & 66.5 & 45.1 & 99.2 & {91.8} & 2.5 & 44.3 \\
		DINOv2, ViT-g, 16x16 patch size, Mask2Former decoder & \textbf{78.2} & 23.2 & 99.2 & 87.9 & {5.0} & {63.6} \\
		PixOOD w/ ResNet-101 DeepLabv3~\cite{vojivr2024pixood} & 43.2	& \textbf{58.5} &	93.5 &	84.0 &	15.1 &	54.6 \\
		Ensemble C & 73.9 & {47.4} & 99.1 & \textbf{92.5} & 52.7 & \textbf{34.7} \\
		DeiT III (IN1K), ViT-S, 16x16 patch size, linear decoder &  44.9 & 54.7 & 97.9 & 89.2 & \textbf{1.7} & 54.2 \\
            \midrule
		\textit{Baseline}: SegFormer-B5~\cite{xie2021segformer} & 67.4 & 24.1 & 97.2 & 77.2 & {30.7} & {71.9} \\
		\textit{Baseline}: ObsNet-R101-DLv3plus~\cite{besnier2021triggering} & 65.3 & 32.1 & 98.5 & 87.8 & {45.6} & {63.4} \\
		\textit{Baseline}: Mask2Former-SwinB~\cite{cheng2022masked} & 67.2 & 13.2 & 90.4 & 47.0 & {55.1} & {82.8} \\
		\bottomrule
	\end{tabular}}
	
	\caption{Track 1 -- DINOv2-OOD -- Semantic metrics for valid pixel predictions and their confidence, averaged across all subsets except SMIYC, computed for ablated models and other approaches.}
	\label{tbl:dino_semantic}
\end{table}

Another noteworthy observation is that all our models have relatively low ECE values, indicating that they are well-calibrated, even though no explicit calibration techniques were applied. Even the DeiT-III-based model, which scores low on the overall BRAVO score, achieves a low ECE value of 1.7. Therefore, further investigation is needed to understand why the ECE values are so low.

Finally, the results suggest that more accurate models in terms of mIoU tend to be worse at identifying their own errors, as indicated by the AUPR-Error metric. However, they excel at identifying correct predictions, as shown by the AUPR-Success metric. It is possible that this happens simply because errors by accurate models are rarer, making it harder to identify them.

Overall, the results show that mIoU, which does not depend on prediction confidence, does not correlate well with the other metrics that do.

OOD results are detailed in \cref{tbl:dino_ood}. Surprisingly, our configuration with worst confidence estimation for valid pixels, DINOv2 with ViT-g/16 and Mask2Former, achieves the highest AUPRC of 84.1, the highest AUROC of 98.8, and the lowest FPR@95 of 4.5 for detecting invalid pixels. This suggests that the mask classification framework used by Mask2Former, where per-class masks are predicted separately, allows this decoder to more accurately identify which pixels belong to the mask of an in-distribution class and which do not. Additionally, while a smaller patch size results in worse confidence estimation for valid pixels (see~\cref{fig:dinov2_ood}), it helps in identifying invalid pixels. Qualitative analyses show that the smaller patch size enables the model to better separate valid and invalid pixels, as it can capture more fine-grained details.
Finally, while scaling from ViT-L to ViT-g improves mIoU for valid pixels (see~\cref{tbl:dino_semantic}), OOD detection performance shows a noticeable degradation.

\begin{table}[t!]
	\centering
	\resizebox{0.6\linewidth}{!}{
	\begin{tabular}{lccc}
		\toprule
		\textbf{Method} & \textbf{AUPRC $\uparrow$} & \textbf{AUROC $\uparrow$} & \textbf{FPR@95 $\downarrow$} \\
		\midrule
		DINOv2, ViT-L, 8x8 patch size, linear decoder & 81.7 & {97.7} & 12.9 \\
		DINOv2, ViT-L, 16x16 patch size, linear decoder & 76.7 & 97.1 & 15.0 \\
		DINOv2, ViT-g, 16x16 patch size, linear decoder & 74.3 & 96.9 & 15.3 \\
		DINOv2, ViT-B, 16x16 patch size, linear decoder & 70.6 & 96.6 & 15.1 \\
		DINOv2, ViT-S, 16x16 patch size, linear decoder & 58.9 & 94.9 & 20.2 \\
		DINOv2, ViT-g, 16x16 patch size, Mask2Former decoder & \textbf{84.1} & \textbf{98.8} & \textbf{4.5} \\
		DeiT III (IN1K), ViT-S, 16x16 patch size, linear decoder & 30.0 & 86.5 & 38.6 \\
		\bottomrule
	\end{tabular}}
	
	\caption{Track 1 -- DINOv2-OOD -- OOD metrics for detecting OOD objects by identifying invalid pixels based on prediction confidence, averaged over the SMIYC and Synobjs subsets, computed for ablated models.}
	\label{tbl:dino_ood}
\end{table}

Overall, the results indicate that the models best at identifying invalid pixels are not necessarily the same ones that excel at correctly classifying valid pixels or accurately estimating their confidence for valid pixels

\subsection{Track 1: PixOOD -- Czech Technical University in Prague}
\label{sec:pixelood1st}
\vspace{-5pt}
\textit{Authors: Tomáš Vojíř, Jan Šochman and Jiří Matas}
\vspace{3pt}

We describe how the semantic segmentation and the confidence scores are computed for all submitted methods. We also discuss the training details with the focus on differences to the original PixOOD~\cite{vojivr2024pixood}.

\parag{Semantic Segmentation.} The semantic class $c \in \{1, 2, \dots, C\}$ for each pixel $p \in \{(y, x)\}^{H \times W}$ of an image $I \in \mathbb{R}^{H \times W \times 3}$ is computed from logits $l \in \mathbb{R}^{H \times W \times C}$ simply as: $c_p^* = \arg\max_{c} (l_p^c)$. The logits are computed using different decoders in each variant as discussed below.

\parag{Confidence.} The confidence of the semantic segmentation (\textit{i.e.} 1 -- OOD score) in all variants of the PixOOD method is computed as $s_I$ by Eq. (3) from PixOOD~\cite{vojivr2024pixood}. Because of the quantization required for saving the results to a 16-bit PNG format (\textit{i.e.}, into the 65,536 values), the score is re-normalized so that the ``effective range'' of the score is well represented. Since the score is calibrated and directly corresponds to the false positive rate of the in-distribution data (training data) the ``effective range'' is mostly limited to the first 5\% -- \textit{i.e.}, (0.0, 0.05), thus, the re-normalization maps the score piece-wise linearly as follows: $[0.0, 0.05] \rightarrow [0.0, 0.8]$ and $[0.05, 1.0] \rightarrow [0.8, 1.0]$ where the square brackets denote inclusion of the boundary in the range. Note that this re-normalization is only useful if we need to quantize the results for some reason. The calculation of the $s_I$ score in PixOOD is performed on a per-class basis, we report the score associated with the predicted class $c_p^*$.

\parag{Models.} The following three PixOOD variants were submitted to the challenge.

\begin{itemize}[leftmargin=*, label=--]
\item \textit{PixOOD.} This is the default method exactly as described in the PixOOD paper with the checkpoints used from the published codebase.

\item \textit{PixOOD w/ DeepLabv3 Decoder.} This method replaces the simple MLP segmentation head in PixOOD by a more complex DeepLab v3~\cite{chen2018encoder} decoder.
The input to the decoder are concatenated features from layers (17, 23) and (5, 11) of DINOv2~\cite{oquab2023dinov2} encoder for the \textit{ASPP} and \textit{fine} bottleneck layers respectively. The output of the decoder is used to generate the logits for the N-P task and the computation of the OOD and segmentation scores.

\item \textit{PixOOD w/ ResNet-101 DeepLabv3.} This method replaces the simple MLP segmentation head in PixOOD by a complete DeepLab v3 network~\cite{chen2018encoder} with a ResNet-101~\cite{he2016deep} backbone. It takes an image as input instead of the latent representation of the DINOv2~\cite{oquab2023dinov2} image encoder used in the previous two methods. The DeepLab v3 output logits (\textit{i.e.}, output of the last layer of the network before any \texttt{argmax} operation) are used for the N-P task and the computation of the OOD and segmentation scores.
\end{itemize}

\begin{table}[t!]
	\centering
	\resizebox{0.4\linewidth}{!}{
		\begin{tabular}{lccc}
			\toprule
			\textbf{Method} & \textbf{BRAVO$\uparrow$} & \textbf{Semantic$\uparrow$} & \textbf{OOD$\uparrow$} \\
			\midrule
			PixOOD & 53.5 & 40.4 & 79.1 \\
			PixOOD w/ DeepLabv3 Decoder & 59.4 & 46.1 & 83.5 \\
			PixOOD w/ ResNet-101 DeepLabv3 & \textbf{61.2} & \textbf{58.7} & \textbf{64.0} \\
			\bottomrule
	\end{tabular}}
	
	\caption{Track 1 -- PixOOD -- Analysis.}
 \vspace{-8pt}

	\label{tbl:pixood_track1}
\end{table}

\parag{Training.} All variants of the segmentation part of the PixOOD method were trained using the same regime for 30 epochs on the Cityscapes dataset with the learning rate set to 0.0001 using the AdamW optimizer without scheduling the learning rate. The submitted variants used basic image augmentations, \textit{i.e.}, random crop of size 1,792 of the longer side while keeping the aspect ratio and random horizontal flip with probability 0.5. These augmentations were used only during training of the head that produces the logits. The calculation of the Condensation algorithm and the N-P decision strategies were the same for all methods and follows the default settings described in PixOOD~\cite{vojivr2024pixood}.

\subsection{Track 1: Ensemble -- McGill University}
\label{sec:ensemble_track1}
\vspace{-5pt}
\textit{Authors: Michael Smith and Frank Ferrie}
\vspace{3pt}

The solutions to both tracks involve ensembles, albeit in different configurations. For both of them, we use ensembles in a standard configuration where we have $Q$ models~\cite{malinin2019uncertainty}:$\left\{ P(y \mid x^*, \theta^{(q)}) \right\}_{q=1}^{Q}, \quad \theta^{(q)} \sim p(\theta \mid \mathcal{D}).$
Each one of these models is capable of generating a prediction $y$ from a test input $x^*$ with weights $\theta^{(q)}$ constrained by the prior $p(\theta)$. The predictions of these models can be aggregated through the predictive posterior as the mean across models, \textit{i.e.}:$P(y \mid x^*, \mathcal{D}) = \frac{1}{Q} \sum_{q=1}^{Q} P(y \mid x^*, \theta^{(q)}).$ With $P(y \mid x^*, \mathcal{D})$, we now have a confidence assigned to each class, with the maximum across the set of classes providing the predicted class and associated confidence in the prediction as required by the BRAVO challenge.

For Track 1, we address both aspects of our hypothesis. For the first part on model diversity, we chose to use ensembles of two models: Mask2Former~\cite{cheng2022masked} and HMSA (Hierarchical Multi-Scale Attention for Semantic Segmentation)~\cite{tao2020hierarchical}. In all cases for this track, pretrained Cityscapes models available online are used. Note that for the Mask2Former approach, we build off the code and use models provided by the authors of RbA (Rejected By All)~\cite{nayal2023rba}, whose main contribution is a scoring function that takes as input the output of a Mask2Former model.
We explicitly denote the approaches where we use said scoring function as being RbA, but otherwise refer to the approach as Mask2Former. Ideally, we would have used more models, but this was not possible due to time and resource constraints. The two models were chosen as they have very different architectures, and thus are good candidates for exploring model diversity in terms of architectures. They also satisfy the other part of our hypothesis as they are known to perform very well in terms of semantic segmentation performance and out-of-distribution performance, on the Cityscapes~\cite{cordts2016cityscapes} and SMIYC~\cite{chan2021segmentmeifyoucan} leaderboards, providing a good starting point in terms of performance.

\parag{Models.} Ensemble Configuration A \& C involve combining the predictions of Mask2Former~\cite{cheng2022masked} and HMSA~\cite{tao2020hierarchical}. Configuration A consists of two models: the Swin-L model of Mask2Former and the \texttt{nimble-chihuahua} model from HMSA. Configuration C adds one additional model in the form of the Swin-B model for Mask2Former, for a total of three models. We apply the softmax operator to the logits of each model independently, which then gives us $Q = \{2, 3\}$ models, giving us a prediction and associated confidence in each class for every pixel.

\begin{table}[t!]
	\centering
	\resizebox{0.4\linewidth}{!}{
		\begin{tabular}{lccc}
			\toprule
			\textbf{Method} & \textbf{BRAVO$\uparrow$} & \textbf{Semantic$\uparrow$} & \textbf{OOD$\uparrow$} \\
			\midrule
			Ensemble A & 59.9 & 67.3 & 53.9 \\
			Ensemble C & \textbf{61.1} & 64.3 & \textbf{58.2} \\
			HMSA & 36.0 & \textbf{70.6} & 24.2 \\
			\bottomrule
	\end{tabular}}
	
	\caption{Track 1 -- Ensemble -- Analysis.}
 \vspace{-8pt}

	\label{tbl:ensemble_track1}
\end{table}
Given the results presented in \cref{tbl:ensemble_track1}, we can make a few observations. The first is that straightforward ensembles using the mean achieves very respectable results, doing relatively well in combining the divergent semantic and OOD performance of the Mask2Former and HMSA models. It is not quite able however to achieve the best of both worlds, as demonstrated by the greater performance of the RbA and HMSA baselines. It is clear as well that other approaches, such as PixOOD (\cref{sec:pixelood1st}), suffer from some trade-offs in terms of semantic and OOD performance when making architecture changes. In this particular case, it is safe to say that network architecture can play a very significant role in performance, and the performance of ensembles can be heavily influenced by it. Our second observation is that it is clear model diversity can play an important role as well, as the performance of our approach is entirely due to being able to combine two approaches that specialize in different metrics. However, with only two models at play for Track 1, we cannot make any definitive statements.

\subsection{Track 1: PhyFea -- University Of Bologna \& ETH Zurich}
\label{sec:phyfea}
\vspace{-5pt}
\textit{Authors: Shamik Basu, Christos Sakaridis and Luc Van Gool}
\vspace{3pt}

Our approach PhyFea (Physically Feasible Semantic Segmentation) enhances the performance of the baseline segmentation architectures $\phi(X)$ as described in our paper by retraining it with physical priors ``inclusion constraint'' and ``discontinued class'' incorporated by PhyFea. After retraining, we can observe an improvement in the mIOU score during inference by the baseline architectures. For this challenge, the baseline architecture we have taken is Segformer-B4~\cite{xie2021segformer}. The training overview is explained in~\cite{basu2024physically}.

\parag{Semantic segmentation.} The semantic class $c \in \{1, 2, 3, \dots, C\}$ for each pixel $p \in (y, x)^{H \times W}$ of an image $I \in \mathbb{R}^{(3, H, W)}$ is computed from logits $\phi(I) \in \mathbb{R}^{(C, H, W)}$ as: $S^* = \arg\max_{s} (\phi(I)).$

\parag{Confidence.} The confidence of the semantic segmentation (\textit{i.e.}, 1 -- OOD score) in the baseline model $\phi(X)$ is computed on its output $\phi(I)$. First, $\phi(I)$ is bounded as $0 \leq \phi(I) \leq 1$ in order to represent the effective range in a better way. Then, quantization is performed and saved in 16-bit PNG format.

\parag{Training.} In~\cite{basu2024physically}, we show the architectural overview of PhyFea. A 2D semantic segmentation model as baseline network $\phi(X)$ takes an image $\mathbf{I} \in \mathbb{R}^{(3, H, W)}$ as input and produces the raw output $\phi(\mathbf{I}) \in \mathbb{R}^{(C, H, W)}$ where $C$ is the number of classes present in the dataset. PhyFea takes $\phi(\mathbf{I})$ as input and produces an absolute difference of two loss values $l_{opening}$ and $l_{dilation}$ generated by the two operations performed in PhyFea, namely opening and selective dilation. Opening solves the inclusion constraint problem and selective dilation solves the discontinued class problem. The absolute difference $\left| l_{opening} - l_{dilation} \right|$ is then added to the cross-entropy loss (denoted by $l_{cross-entropy}$) of $\phi(X)$ to obtain the total loss. Here $\alpha$ is a hyperparameter and it is used to balance the loss of PhyFea and the baseline network. $l_{total-loss}$ is backpropagated to optimize the weights of the baseline network by obtaining the \texttt{argmin} of $l_{total-loss}$ as: $l_{total-loss} = l_{cross-entropy} + \alpha * \left| l_{opening} - l_{dilation} \right|, \quad 0 < \alpha < 1 $ and $S^* = \arg\min_s (l_{total-loss})$. PhyFea is end-to-end differentiable in order to incorporate the physical priors (\textit{i.e.}, inclusion constraint and discontinued class) while re-training the baseline network and it is free of any parameterized component like convolution kernel or MLP.

\subsection{Track 2: InternImage-OOD -- CASIA \& Objecteye}
\label{sec:internimage_ood}
\vspace{-5pt}
\textit{Authors: Long Qian, Bingke Zhu, Yingying Chen, Ming Tang, and Jinqiao Wang}
\vspace{3pt}

We introduce InternImage-OOD, which integrate the power of general large-scale vision foundation model and the efficiency of simple clustering algorithm, to solve the challenge. We fine-tuned the vision foundation model to enhance its semantic segmentation capabilities, and integrated a simple clustering algorithm to improve the model's OOD detection performance, all while maintaining efficiency. We employ K-Means clustering as a post-processing technique to improve OOD detection followed by PixOOD~\cite{vojivr2024pixood} (\cref{sec:pixelood1st} and~\cref{sec:pixelood2nd}). Our method consists of two stages. First, the input image is processed using the InternImage model to perform basic semantic
segmentation, resulting in the predicted class map $P$, the confidence map $C$, and the image feature embeddings $F$. Then, we apply a K-Means-based OOD detection method for post-processing the results.

As the core of the pipeline, we apply the InternImage to perform basic semantic segmentation. $P = \left\{ p_i \in \mathbb{R}^{H \times W} \right\}_{i=1}^{K} = \text{InternImage}(I)$, $C = \left\{ c_j \in \mathbb{R}^{H \times W} \right\}_{j=1}^{K} = \text{Confidence}(I)$ and $F = \left\{ f_k \in \mathbb{R}^{D \times H \times W} \right\}_{k=1}^{M} = \text{FeatureEmbedding}(I)$ where: $P$ is the set of predicted pixel-wise class labels from the InternImage model, $C$ is the set of confidence maps associated with each class prediction, $F$ is the set of feature embeddings extracted from the feature pyramid, $H$ and $W$ represent the height and width of the image, respectively, $K$ denotes the number of classes, $D$ is the dimensionality of the feature embedding for each pixel, and $M$ represents the number of feature levels in the feature pyramid.

In the second stage, K-Means clustering is applied to the feature embeddings for OOD detection and further refinement, with $OODMask = \text{K-Means}(F_k)$ and $UpdatedConfidence(C) = \text{UpdateMask}(OODMask, C)$. $OODMask$ is the mask, which shows the region of OOD, obtained from K-Means clustering applied to feature embeddings. UpdateMask is the function that updates the confidence map $C$ based on the OOD mask.

Here, \textit{feature embeddings} are taken from different depths of the model, representing the model from shallow to deep features. For each class in the \textit{predicted class map} $P$, where regions are recognized as low-confidence regions, the corresponding \textit{feature embeddings} $F$ are passed to the trained KMeans models. By calculating the distances between the embeddings and the cluster centroids, OOD regions are identified based on how far they deviate from the known clusters. Regions with distances significantly higher than the average are marked as OOD, and the corresponding areas in the \textit{confidence map} $C$ are updated with lower values to reflect the uncertainty.

\parag{Datasets.} Considering the time constraints of the competition and in order to demonstrate the advantages of this method to still achieve excellent performance on a small number of datasets, we only used Mapillary Vistas and Cityscapes datasets to complete this experiment, which have about 50,000 images in total.

\parag{Implementation Details.} We choose the \textit{Upernet\_InternImage\_XL}~\cite{wang2022internimage} as our backbone, and train the model on Mapillary Vistas and Cityscapes datasets one by one, during which we choose the AdamW optimizer with a learning rate of 2e-5, a batch size of 8, and a weight decay of 0.05. Besides, we use K-Means as our optimization method on OOD, in which we only use a few images and a very small number of the cluster centroids, considering the limited time.

\begin{table}[t!]
	\centering
	\resizebox{0.6\linewidth}{!}{
		\begin{tabular}{lccc}
			\toprule
			\textbf{Method} & \textbf{BRAVO$\uparrow$} & \textbf{Semantic$\uparrow$} & \textbf{OOD$\uparrow$} \\
			\midrule
			InternImage & 62.1 & 69.3 & 56.2 \\
			+ KMeans-Based OOD & \textbf{62.6} & \textbf{69.3} & \textbf{57.1} \\
			\bottomrule
	\end{tabular}}
	
	\caption{Track 2 -- InternImage-OOD -- Ablation Study.}
 \vspace{-8pt}

	\label{tbl:internimage_abl}
\end{table}

\parag{Results.} Our solution achieved BRAVO-Index of 62.6 (see~\cref{tbl:track2_all}), and ranked 1st in the Multi-domain training Track. To show the effectiveness of our solution, we conduct an ablation study. As show in \cref{tbl:internimage_abl}, we achieved a modest improvement with a very small amount of data and clustering centers.

\subsection{Track 2: PixOOD -- Czech Technical University in Praque}
\label{sec:pixelood2nd}
\vspace{-5pt}
\textit{Authors: Tomáš Vojíř, Jan Šochman and Jiří Matas}
\vspace{3pt}

We refer to \cref{sec:pixelood1st} for details of the method and different model variants. For Track 2, only the \textit{PixOOD w/ DeepLabv3 Decoder} variant is submitted (see~\cref{tbl:track2_all}). The method was trained on the combined Cityscapes and BDD100K datasets. The BDD100K data set was randomly sub-sampled such that the number of images is roughly equal (taking 1/3 of the data) to the size of Cityscapes. As the resolution of these two datasets is different, a smaller random crop of size 1036 (of the longer side while keeping the aspect ratio) was used during training.

\subsection{Track 2: Ensemble -- McGill University}
\label{sec:ensemble_track2}
\vspace{-5pt}
\textit{Authors: Michael Smith and Frank Ferrie}
\vspace{3pt}

We refer to \cref{sec:ensemble_track1} for the common theory. Here we present the methodology differences  adopted for Track 2 and the corresponding results.

For Track 2, our primary goal was to evaluate the potential use of different datasets as a source of model diversity for the ensembles. The BRAVO challenge is set up to evaluate only the standard 19 Cityscapes evaluation classes, and Track 2 allows for the use of multiple datasets. With all of these datasets placing a clear focus on autonomous driving in some way, they are all formatted to either use the aforementioned 19 Cityscapes classes or use classes similar enough such that they can be mapped to the Cityscapes ones. This presents an opportunity where we can train models on different datasets with different characteristics (including some synthetic ones) while maintaining compatibility with one another. Here, we use the same models as Track 1, with a particular focus on HMSA as we train a model with that architecture for each allowed dataset.

\parag{Training.} Before we could evaluate any models, we first needed to obtain one trained model per dataset with the HMSA architecture~\cite{tao2020hierarchical}. The models for each were generated as follows:

\begin{itemize}[leftmargin=*, label=--]
	\item \textit{Cityscapes}: We used the author-provided \texttt{nimble-chihuahua} model~\cite{tao2020hierarchical}.
	
	\item \textit{Mapillary}: After converting the dataset to use Cityscapes labels, we trained the model using transfer learning from the \texttt{fast-rattlesnake} model~\cite{tao2020hierarchical} as provided by the authors and with the same training configuration as they provide with the code for their Mapillary model, except the number of epochs, which we set to 15 as we needed to adapt the model to the new class scheme.
	
	\item \textit{GTA5}: We first removed some corrupted images and then resized all images and ground truth masks to (1914, 1052). The model was then trained with transfer learning from the \texttt{fast-rattlesnake} model, with the same training settings as used by~\cite{tao2020hierarchical} for their \texttt{cityscapes\_sota} training configuration.
	
	\item \textit{SHIFT}: We used the dataset author-provided label mapping to Cityscapes and trained the model with transfer learning from the \texttt{outstanding-turtle} model from~\cite{tao2020hierarchical}. The training parameters are the same as with the GTA5 model, except for the learning rate, which is set to 0.005.
	
    \item \textit{BDD100K}: This model is trained via transfer learning from the \texttt{industrious-chicken} model~\cite{tao2020hierarchical} with the same settings as the GTA5 model.

	\item \textit{IDD}: This transfer learning source for this model is the \texttt{outstanding-turtle} model. Parameters are the same as the GTA5 model.
	
	\item \textit{Wild Dash 2}: The \texttt{nimble-chihuahua} model is used for pretraining in this case, with parameters the same as the SHIFT model.
\end{itemize}

\begin{table}[t!]
	\centering
	\resizebox{0.4\linewidth}{!}{
		\begin{tabular}{lccc}
			\toprule
			\textbf{Method} & \textbf{BRAVO$\uparrow$} & \textbf{Semantic$\uparrow$} & \textbf{OOD$\uparrow$} \\
			\midrule
			Ensemble A & 40.6 & 66.0 & 29.4 \\
			Ensemble B & 45.5 & 64.6 & 35.2 \\
			Ensemble C & 58.8 & 64.5 & 54.0 \\
			\bottomrule
	\end{tabular}}
	
	\caption{Track 2 -- Ensemble -- Analysis.}
	\label{tbl:ensemble_track2}
\vspace{-8pt}
 
\end{table}

\cref{tbl:track2_all} and~\cref{tbl:ensemble_track2} reports results in Track 2.
We can see that while using several models helps the BRAVO score for Multi-dataset ensemble configuration A over the HMSA baseline, it does so by improving the OOD score at the expense of the semantic score.
Multi-dataset configurations B and C, however, show that adding the two Mask2Former models, with their ability to do better at OOD detection, is much more impactful than using more models of the same HMSA architecture trained on different datasets. More generally, neither of the two approaches tested on both Track 1 and 2 (our approach with Ensembles and PixOOD) show any notable improvement from using more datasets.

\section*{Acknowledgements}
We extend our heartfelt gratitude to the authors of ACDC~\cite{sakaridis2021acdc}, SegmentMeIfYouCan~\cite{chan2021segmentmeifyoucan}, and Out-of-context Cityscapes~\cite{franchi2021robust} for generously permitting us to repurpose their benchmarking data. We are also thankful to the authors of GuidedDisent~\cite{pizzati2023physics}, Flare Removal~\cite{wu2021train}, and GenVal~\cite{loiseau2024reliability} for providing the excellent toolboxes that helped synthesize realistic-looking raindrops, light flares, and inpainted objects. All 
% those people 
have collectively contributed to creating BRAVO, a unified benchmark for robustness in autonomous driving.

The BRAVO Challenge is an initiative within ELSA — European Lighthouse on Secure and Safe AI, a network of excellence funded by the European Union. This work was supported by ELSA and was funded by the European Union under grant agreement No.~101070617.

\appendix
\noindent We provide in this document additional details of the BRAVO challenge and of the solutions.

\section{BRAVO Challenge}
\label{sec:bravo_challenge_supp}
\subsection{General Rules}
For the BRAVO Challenge 2024 challenge, the following rules applied:
\begin{enumerate}[label=\alph*)]
    \item The task is semantic segmentation, with pixel-wise evaluation performed on the 19 semantic classes of Cityscapes.
    \item Models in each track must be trained using only the datasets allowed for that track.
    \item Employing generative models for synthetic data augmentation is strictly forbidden.
    \item All results must be reproducible. Participants must submit a white paper containing comprehensive technical details alongside their results. \item Participants must make models and inference code accessible.
    \item Evaluation considers the 19 classes of Cityscapes: `road', `sidewalk', `building', `wall', `fence', `pole', `traffic light', `traffic sign', `vegetation', `terrain', `sky', `person', `rider', `car', `truck', `bus', `train', `motorcycle' and `bicycle'.
    \item Teams must register a single account for submitting to the evaluation server. An organization (e.g. a University) may have several teams with independent accounts only if the teams are not cooperating.
\end{enumerate}

\subsection{Submissions}
For each input image, two files were required: one for the semantic predictions and one for the confidence values. 

The class prediction file must be in PNG format, 8-bit grayscale, with each pixel assigned a value from 0 to 19, representing the 19 classes of Cityscapes. The confidence file must also be in PNG format, but 16-bit grayscale, with each pixel's value ranging from 0 to 65,535, representing the confidence level of the predicted class. Confidence values are evaluated across the entire subset of the dataset simultaneously and, therefore, should be comparable across all images in the subset.
Each prediction and confidence file must have the exact same dimensions as the corresponding input image. Evaluation is performed on a pixel-by-pixel basis.

\section{Submissions digest}
\label{sec:submission_digest_supp}
\begin{figure}[t!]
  \centering
  \includegraphics[width=0.8\textwidth]{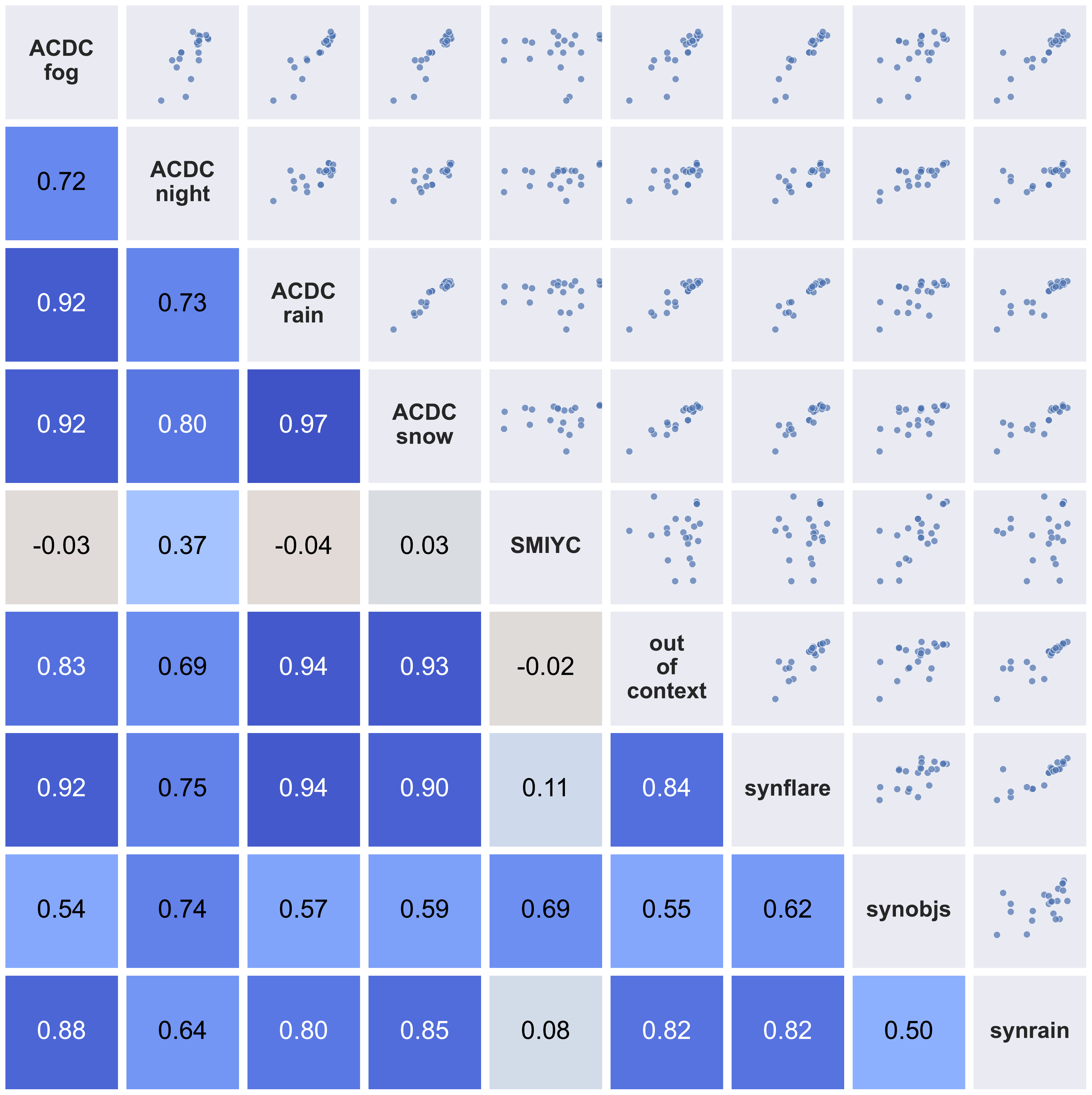}
  \caption{Analysis showing the correlation of the summary metric of each BRAVO subset.}
  \label{fig:correlogram}
\end{figure}

From the correlogram in~\cref{fig:correlogram}, we observed varying degrees of correlation among the metrics aggregated by the BRAVO subsets.

\subsection{Track 1: DINOv2-OOD -- Eindhoven University of Technology}
\label{sec:dinov2ood_sup}
\textit{Authors: Tommie Kerssies, Daan de Geus, Gijs Dubbelman}

\begin{figure}[t!]
  \centering
  \includegraphics[width=0.8\textwidth]{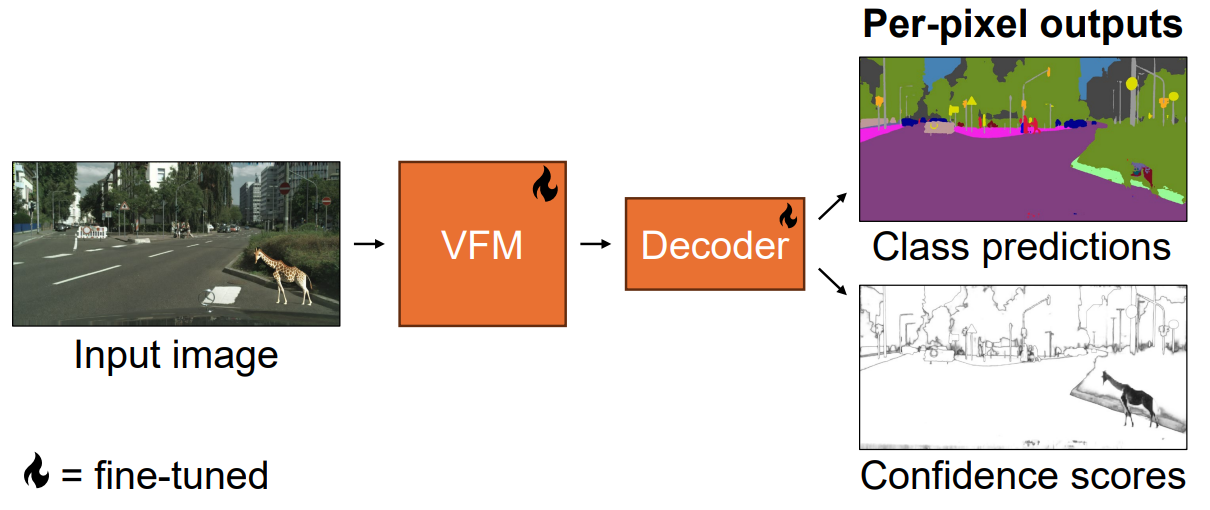}
  \caption{DINOv2-OOD Meta-approach. We take a pre-trained Vision Foundation Model (VFM), attach a simple segmentation decoder, and fine-tune the entire model for semantic segmentation. The segmentation decoder outputs both the per-pixel classification predictions and the associated confidence scores.}
  \label{fig:dinov2_ood}
\end{figure}

\parag{Method.} \autoref{fig:dinov2_ood} overviews the DINOv2-OOD approach.

\parag{Implementation details.} We use the following models from the \texttt{timm} library~\cite{rw2019timm} to initialize the VFM:
\begin{itemize}[leftmargin=*, label=--]
	\item \texttt{deit3\_small\_patch16\_224.fb\_in1k};
	\item \texttt{vit\_small\_patch14\_dinov2};
	\item \texttt{vit\_base\_patch14\_dinov2};
	\item \texttt{vit\_large\_patch14\_dinov2};
	\item \texttt{vit\_giant\_patch14\_dinov2}.
\end{itemize}

The models are fine-tuned for 40 epochs using two A6000 GPUs, with a batch size of 1 per GPU and gradient accumulation over 8 steps, resulting in an effective batch size of 16. Our implementation follows the details provided in~\cite{kerssies2024benchmark}. Notably, the learning rate for the VFM weights is set to be 10$\times$ smaller than the overall learning rate, as this configuration empirically yields better results. For the Mask2Former decoder, we employ a variant specifically adapted for use with a single-scale ViT encoder, as introduced in~\cite{kerssies2024benchmark}.

\begin{table}[h!]
	\centering
	\resizebox{1.0\linewidth}{!}{
	\begin{tabular}{lcccccc}
		\toprule
		\textbf{Method} & \textbf{ACDC $\uparrow$} & \textbf{SMIYC $\uparrow$} & \textbf{Out-of-context $\uparrow$} & \textbf{Synflare $\uparrow$} & \textbf{Synobjs $\uparrow$} & \textbf{Synrain $\uparrow$} \\
		\midrule
		DINOv2, ViT-L, 8x8 patch size, linear decoder & 67.3 & 89.9 & 71.0 & 72.7 & \textbf{76.7} & \textbf{73.9} \\
		DINOv2, ViT-L, 16x16 patch size, linear decoder & \textbf{69.4} & 89.3 & 70.4 & 72.4 & 75.1 & 73.8 \\
		DINOv2, ViT-g, 16x16 patch size, linear decoder & 67.7 & {88.2} & 71.0 & \textbf{73.2} & 74.7 & 73.2 \\
		DINOv2, ViT-B, 16x16 patch size, linear decoder & 68.5 & 87.9 & \textbf{71.2} & 72.8 & 74.0 & 73.0 \\
		DINOv2, ViT-S, 16x16 patch size, linear decoder & 66.9 & 83.1 & {70.2} & 70.6 & 68.6 & 72.9 \\
		DINOv2, ViT-g, 16x16 patch size, Mask2Former decoder & 49.1 & \textbf{94.4} & 40.9 & 53.9 & 64.3 & 60.1 \\
		DeiT III (IN1K), ViT-S, 16x16 patch size, linear decoder & 58.8 & 50.1 & 65.1 & 71.4 & 58.5 & 65.6 \\
		\bottomrule
	\end{tabular}}
	
	\caption{Track 1 -- DINOv2-OOD -- Harmonic means of semantic and OOD metrics for each subset in the BRAVO benchmark dataset, computed for ablated models.}
\end{table}

\subsection{Track 1: PixOOD -- Czech Technical University in Prague}
\label{sec:pixelood1st_sup}
\vspace{-5pt}
\textit{Authors: Tomáš Vojíř, Jan Šochman and Jiří Matas}
\vspace{3pt}

\begin{figure}[t!]
  \centering
  \includegraphics[width=0.8\textwidth]{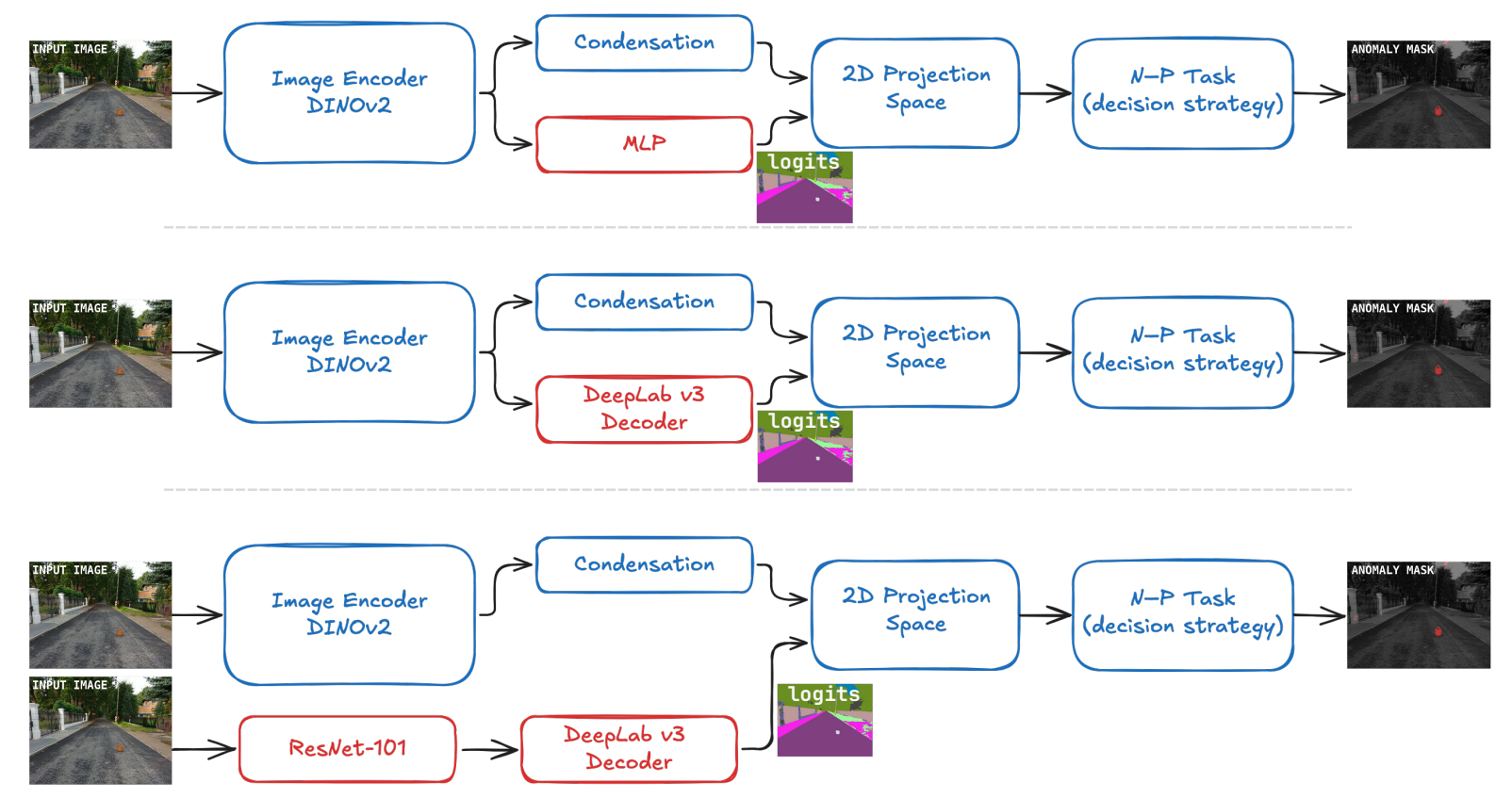}
  \caption{PixOOD Variants. Simplified block representation of the PixOOD framework for different submitted variants. From top to bottom: PixOOD, PixOOD w/ DeepLab Decoder and PixOOD w/ ResNet101 DeepLab. The blue color denotes blocks that are the same for all variants and are described in the PixOOD. The red color denotes the differences between the methods in the semantic segmentation branches.}
  \label{fig:pixood}
\end{figure}

\parag{Method.} \autoref{fig:pixood} visualizes the three PixOOD variants that were submitted to the BRAVO challenge 2024.

\subsection{Track 1: PhyFea -- University Of Bologna \& ETH Zurich}
\label{sec:phyfea_supp}
\vspace{-5pt}
\textit{Authors: Shamik Basu, Christos Sakaridis and Luc Van Gool}
\vspace{3pt}

\begin{figure}[t!]
  \centering
  \includegraphics[width=0.8\textwidth]{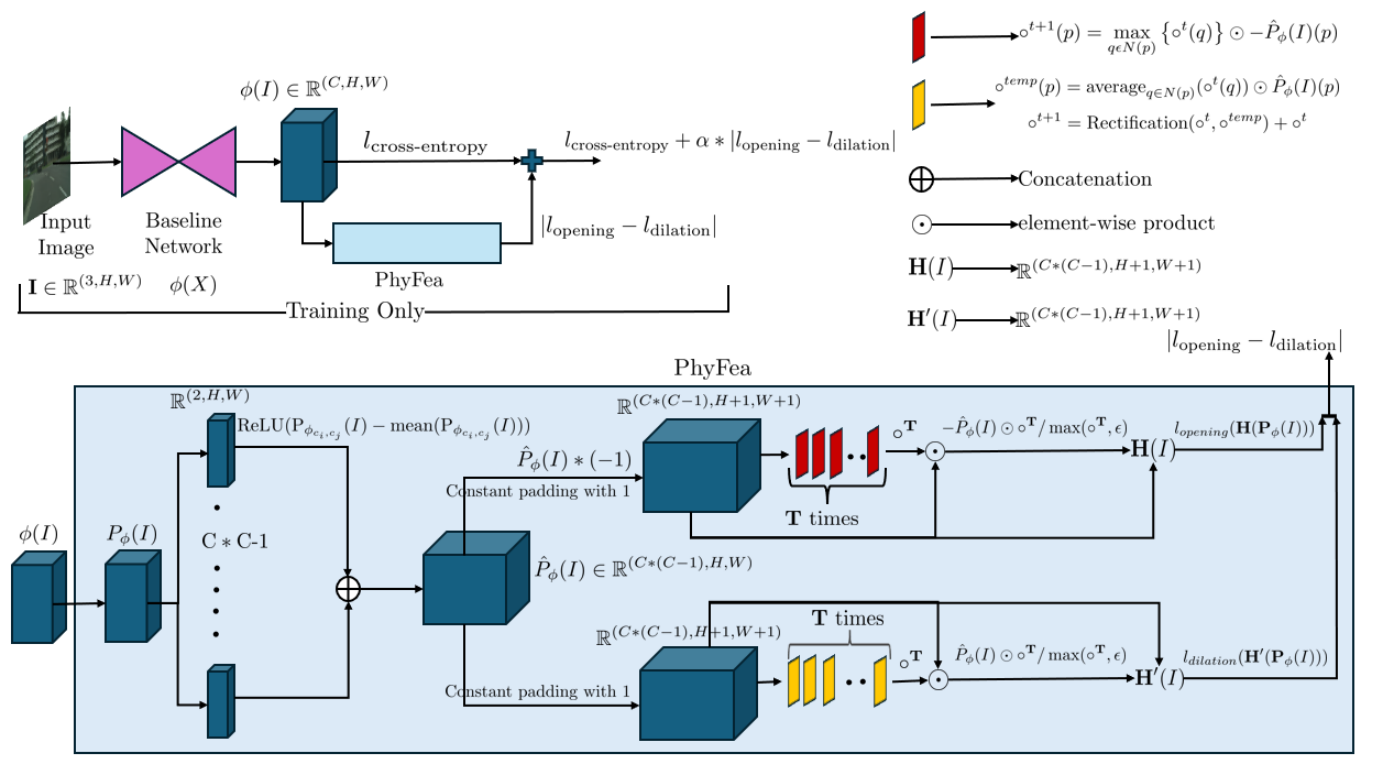}
  \caption{PhyFea approach. Top left: illustration of the complete network architecture, where the cross-entropy loss of the baseline network is added to the losses of PhyFea. Bottom: the pipeline of PhyFea, where red-colored boxes are iterations for opening and yellow colored boxes are for selective dilation. Top right: legends for various components of PhyFea, such as the operations we apply in iterative manner for area opening and for selective dilation and the two functions to calculate the losses.}
  \label{fig:phyfea}
\end{figure}

\parag{Method.} \autoref{fig:phyfea} overviews the PhyFea approach.

\subsection{Track 2: InternImage-OOD -- CASIA \& Objecteye}
\label{sec:internimage_ood_sup}
\vspace{-5pt}
\textit{Authors: Long Qian, Bingke Zhu, Yingying Chen, Ming Tang, and Jinqiao Wang}
\vspace{3pt}

\begin{figure}[t!]
  \centering
  \includegraphics[width=0.8\textwidth]{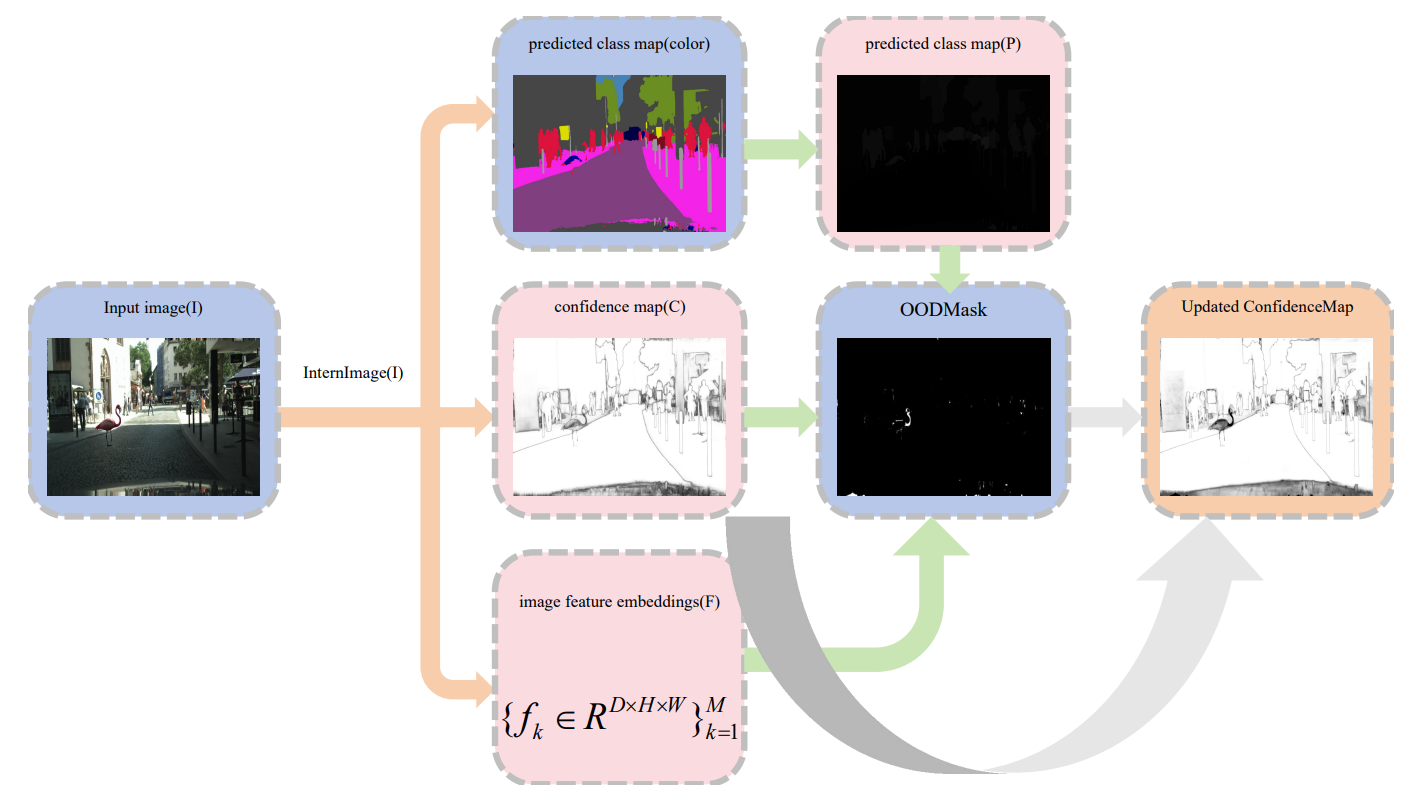}
  \caption{InternImage-OOD. The diagram illustrates the process of OOD detection and confidence map refinement. Starting with an input image $I$, the InternImage model generates both the predicted class map $P$ and the confidence map $C$. Image feature embeddings $F$ are extracted, and K-Means clustering is applied to detect OOD regions, forming the $OODMask$. Finally, the $OODMask$ is used to update the confidence map, resulting in the updated confidence map for further refinement.}
  \label{fig:internimage_ood}
\end{figure}

\parag{Method.} \autoref{fig:internimage_ood} overviews the InternImage-OOD solution.

\subsection{Track 2: Ensemble -- McGill University}
\label{sec:ensemble_track2_sup}
\textit{Authors: Michael Smith and Frank Ferrie}

Below are a number of settings which we set when training all models on each dataset but do not explicitly enumerate in the interest of brevity:

\begin{itemize}[leftmargin=*, label=--]
	\item \textbf{Splits}: In all cases, we used dataset author-provided training splits.
	
	\item \textbf{Tile size}: The training process uses tiling during training to try and ensure a better class distribution when training for some classes that are more rare. We try and set this such that each image can be decomposed into two tiles as best as possible, depending on the size(s) of images contained in the dataset.
	
	\item \textbf{Crop size/resizing}: We set the image resizing and cropping to match the image size(s) in the dataset as best as possible so as to minimize data loss.
	
	\item \textbf{Epochs}: All models are trained for 175 epochs unless otherwise indicated. However, the final model used was the one which achieved the best results on the respective validation set of each dataset during the entire training run and thus may have been trained for fewer epochs.
	
	\item \textbf{Batch sizes}: Training and validation batch sizes are set as high as possible given the VRAM capacity of the GPUs being used for training.
\end{itemize}

\bibliographystyle{splncs04}
\bibliography{main}
\end{document}